\documentclass[lettersize,journal]{IEEEtran}
\usepackage{amsmath,amsfonts}
\usepackage{algorithmic}
\usepackage{algorithm}
\usepackage{array}
\usepackage[caption=false,font=normalsize,labelfont=sf,textfont=sf]{subfig}
\usepackage{textcomp}
\usepackage{stfloats}
\usepackage{url}
\usepackage{verbatim}
\usepackage{graphicx}
\usepackage{cite}

\usepackage{multirow} 
\usepackage{makecell}
\usepackage{pifont}
\usepackage{times}
\usepackage{soul}
\usepackage{url}
\usepackage[hidelinks]{hyperref}
\usepackage[utf8]{inputenc}
\usepackage[small]{caption}
\usepackage{graphicx}
\usepackage{amsmath}
\usepackage{amsthm}
\usepackage{booktabs}
\usepackage{algorithm}
\usepackage{algorithmic}
\usepackage[switch]{lineno}

\begin{document}

\title{SPECIAL: Zero-Shot Hyperspectral Image Classification with CLIP}

\author{
Li Pang,
\and
Jing Yao,
\and
Kaiyu Li,
\and
Jun Zhou,~\IEEEmembership{Fellow,~IEEE,}
Deyu Meng,
Xiangyong Cao
\thanks{Li Pang is with the School of Mathematics and Statistics and the Ministry of Education Key Laboratory of Intelligent Networks and Network Security, Xi’an Jiaotong University, Xi’an, Shaanxi 710049, China (e-mail: 2195112306@stu.xjtu.edu.cn).}
\thanks{Jing Yao is with the Aerospace Information Research Institute, Chinese Academy of Sciences, Beijing, China (e-mail: jasonyao92@gmail.com).}
\thanks{Kaiyu Li and Xiangyong Cao are with the School of Computer Science and Technology and the Ministry of Education Key Laboratory for Intelligent Networks and Network Security, Xi’an Jiaotong University, Xi’an 710049, China (e-mail: likyoo.ai@gmail.com; caoxiangyong@xjtu.edu.cn).}
\thanks{Jun Zhou is with the School of Information and Communication Technology, Griffith University, Nathan, QLD 4111, Australia (e-mail: jun.zhou@griffith.edu.au).}
\thanks{Deyu Meng is with the School of Mathematics and Statistics and the Ministry of Education Key Laboratory of Intelligent Networks and Network Security, Xi’an Jiaotong University, Xi’an, Shaanxi 710049, China, and also with Pazhou Laboratory (Huangpu), Guangzhou, Guangdong 510555, China (e-mail: dymeng@mail.xjtu.edu.cn).}
}


\markboth{Journal of \LaTeX\ Class Files,~Vol.~14, No.~8, August~2021}%
{Shell \MakeLowercase{\textit{et al.}}: A Sample Article Using IEEEtran.cls for IEEE Journals}


\maketitle

\begin{abstract}
Hyperspectral image (HSI) classification aims to categorize each pixel in an HSI into a specific land cover class, which is crucial for applications such as remote sensing, environmental monitoring, and agriculture. Although deep learning-based HSI classification methods have achieved significant advancements, existing methods still rely on manually labeled data for training, which is both time-consuming and labor-intensive.
To address this limitation, we introduce a novel zero-\underline{S}hot hypers\underline{PEC}tral \underline{I}mage cl\underline{A}ssification framework based on C\underline{L}IP (SPECIAL), aiming to eliminate the need for manual annotations. The SPECIAL framework consists of two main stages: (1) CLIP-based pseudo-label generation, and (2) noisy label learning. In the first stage, HSI is spectrally interpolated to produce RGB bands. These bands are subsequently classified using CLIP, resulting in noisy pseudo-labels that are accompanied by confidence scores.
To improve the quality of these labels, we propose a scaling strategy that fuses predictions from multiple spatial scales. 
In the second stage, spectral information and a label refinement technique are incorporated to mitigate label noise and further enhance classification accuracy. 
Experimental results on three benchmark datasets demonstrate that our SPECIAL outperforms existing methods in zero-shot HSI classification, showing its potential for more practical applications. 
The code is available at \url{https://github.com/LiPang/SPECIAL}.
\end{abstract}

\begin{IEEEkeywords}
Hyperspectral Image, Classification, Remote Sensing
\end{IEEEkeywords}

\section{Introduction}
Hyperspectral imaging captures detailed spectral information at multiple wavelengths for each pixel, extending far beyond the capabilities of traditional optical imaging modes.
The rich spectral information in the acquired hyperspectral images (HSIs) enables the identification of materials through their unique spectral signatures, leading to various applications such as mineral exploration~\cite{bishop2011hyperspectral, guha2020mineral,peyghambari2021hyperspectral}, environmental monitoring~\cite{zhang2012application, rajabi2024hyperspectral}, and land cover and land use classification~\cite{yao2022semi, lou2025land}.

Among these applications, HSI classification, which involves assigning each pixel to a specific land cover class, has become a popular research area in recent years. Earlier methods often involve manual feature extraction and traditional machine learning to identify relevant features from spectral data~\cite{tong2013urban,melgani2004classification}. However, these methods suffer from the biased prior knowledge and lack the ability to capture nonlinear relationships under complex scenes. With the significant progress in deep learning techniques, numerous studies have explored its application to HSI classification. Convolution neural networks (CNNs)~\cite{lee2017going,li2019deep,ge2020hyperspectral} and Transformers~\cite{qing2021improved, yao2023extended,zhao2024hyperspectral} have been widely employed to enhance the classification performance. More recently, Mamba~\cite{gu2023mamba} architecture, which benefits from both high computational efficiency and long-range modeling capabilities, has also gained popularity in hyperspectral classification~\cite{yao2024spectralmamba,10604894,he2024igroupss}. Although these models have demonstrated impressive performance, their effectiveness heavily depends on the completeness and accuracy of manually labeled data, which is labor-intensive to obtain. Moreover, due to the diverse spectral characteristics across different sensors and various ground object types, re-annotation of data and labels is often necessary in open scenarios, increasing the pressure on labor demands.

Recently, CLIP~\cite{radford2021learning}, a deep model trained to align images and text in a shared embedding space, has gained increasing popularity in open-vocabulary semantic segmentation~\cite{liang2023open,zhou2023zegclip,lin2023clip}. In these approaches, compact image and text features are generally first obtained through the visual and text encoders, respectively. Then, the image features are upsampled to the original spatial size, allowing each pixel to be classified by measuring the similarity between its features and a set of text features. Several studies have also attempted to apply the CLIP model to semantic segmentation in remote sensing~\cite{zhang2024segclip,li2024segearth}, demonstrating promising performance in open-vocabulary settings. Very recently, DiffCLIP~\cite{zhang2024diffclip} employs CLIP for few-shot hyperspectral classification. However, the application of the CLIP model for zero-shot HSI classification remains unexplored.

To further explore the potential of visual and language models for intelligent hyperspectral interpretation in real-world scenarios, in this article, we propose a novel zero-\textbf{S}hot hypers\textbf{PEC}tral \textbf{I}mage cl\textbf{A}ssification framework based on C\textbf{L}IP (\textbf{SPECIAL}). 
For hyperspectral data in new scenes, we first spectrally interpolate the original HSI to simulate RGB data that is compatible with on-the-shelf CLIP, and then utilize the latter to perform unsupervised classification, obtaining pseudo-labels and corresponding confidence scores for each pixel.
We also introduce the multi-scale mechanism in this process to improve the quality of the pseudo-labels generated by CLIP.
Next, we consider incorporating spectral information to boost the classification performance further. Specifically, we leverage the pseudo-labels provided by CLIP to guide the training of an HSI classification network. Moreover, considering that there are substantial noise in the pseudo-labels, we propose a label refinement strategy that utilizes Gaussian Mixture Models (GMMs). By identifying high-confidence and low-confidence samples and modeling their distributions, we are able to generate soft labels of additional training samples that can effectively refine the training process. Experiments on three datasets demonstrate that our approach is superior to existing unsupervised classification approaches.
To summarize, the main contributions of this article are as follows.
\begin{enumerate}
\item We propose a novel zero-shot HSI classification framework, termed as \textbf{SPECIAL}, which enables HSI classification without the need for manually annotated labels. To the best of our knowledge, \textbf{SPECIAL} is the first zero-shot HSI classification method based on CLIP.
\item To improve the recognition accuracy of CLIP models for objects of varying sizes, a resolution scaling (RS) strategy, which fuses predictions under different image resolutions, is proposed to improve the quality of the pseudo-labels provided by CLIP. 
\item To further improve the performance of HSI classification, we propose a noise-robust framework that incorporates both hyperspectral information and CLIP prior, in which three training subsets are dynamically sampled to alleviate the issue of label noise. 
\end{enumerate}

The structure of this paper is structured as follows: Section \ref{sec:related work} reviews related work on hyperspectral image classification and CLIP-based segmentation. Section \ref{sec:method} details our proposed SPECIAL framework, including pseudo-label generation and noisy label learning. Section \ref{sec:exp} presents extensive experimental results and ablation studies on three benchmark datasets. Finally, Section \ref{sec:conclusion} concludes the paper and discusses future work directions.

\section{Related Work}\label{sec:related work}
\subsection{Hyperspectral Image Classification}
Hyperspectral image (HSI) classification methods are diverse and evolving, spanning from traditional machine learning approaches to modern deep learning methods. Early machine learning (ML)-based methods relied on manually crafted features from the spectral signatures and then applied classical classifiers. For example, support vector machines (SVM)~\cite{gualtieri2000support, melgani2004classification,tong2013urban} and random forests (RF) \cite{ham2005investigation, xia2017random, amini2018object} were a popular choice for HSI classification due to their effectiveness in high-dimensional space. These traditional methods often incorporated dimensionality reduction (e.g., PCA) \cite{rodarmel2002principal, farrell2005impact} to handle the high spectral dimensionality. While such approaches are straightforward and require little data, they generally struggle to capture complex nonlinear relationships and spatial context in the imagery.

In the past decade, deep learning (DL) methods have dramatically advanced HSI classification performance. Convolutional neural networks (CNNs) were among the first DL models applied to HSI, initially processing the data as either 1-D spectral sequences or 2-D spatial images~\cite{li2019deep,ge2020hyperspectral}. Beyond CNNs, Transformer-based models have entered HSI analysis recently. They treat the HSI as a sequence of tokens (patches or pixels) and use self-attention to model long-range dependencies. Yao et al.~\cite{yao2023extended} introduced a vision transformer for land-cover classification that integrates spectral and spatial tokens, and Zhao et al.~\cite{zhao2024hyperspectral} applied a transformer with separable convolutions to efficiently handle HSI data. These transformer models can capture global context better than local CNN filters, albeit with higher data requirements. Another emerging approach is the Mamba architecture, based on state-space models, which achieves linear time complexity with the ability to model long-range interactions. Mamba-based HSI classifiers~\cite{yao2024spectralmamba,10604894,he2024igroupss} have shown promising results. They effectively encode the HSI as a sequence and use learned state transitions to capture global context, offering a memory-efficient alternative to Transformers. Overall, deep learning methods significantly outperform traditional methods by automatically learning expressive spectral-spatial features. However, they usually demand a considerable amount of labeled data for training and can struggle when labels are scarce or when facing new sensor types or acquisition conditions.

To mitigate the reliance on labeled data, researchers have explored pretraining and domain-adaptive techniques for HSI classification. Self-supervised representation learning and pretraining on auxiliary tasks have been used to initialize models that can then be fine-tuned with few labels. For example, Lee et al.~\cite{lee2022exploring} pre-trained models on a large corpus of unlabeled HSI data (and even cross-domain data) to learn general features before fine-tuning on a target HSI classification task. Contrastive learning methods have also been employed: Guan and Lam~\cite{guan2022cross} propose a cross-domain contrastive learning framework that learns sensor-invariant features by pulling together representations of the same pixel under different augmentations and pushing apart others. Li et al.~\cite{li2023supervised} combine supervised contrastive learning with unsupervised domain adaptation to transfer a model from a source domain to a target domain with minimal labels. These approaches improve generalization to new scenes or sensors but typically still assume at least some labeled data in the target domain for fine-tuning or validation. Despite these advances, all the above methods still require some human annotations. In scenarios where obtaining labels is extremely difficult (e.g., new hyperspectral sensors or exotic environments), it becomes attractive to explore zero-shot learning approaches, which we discuss later. This has motivated our work in eliminating the need for any task-specific labeled data by leveraging prior knowledge from a pretrained vision-language model \cite{sun2025mask,chen2025cangling}.

\begin{figure*}[h]
  \centering
  \includegraphics[width=1\textwidth]{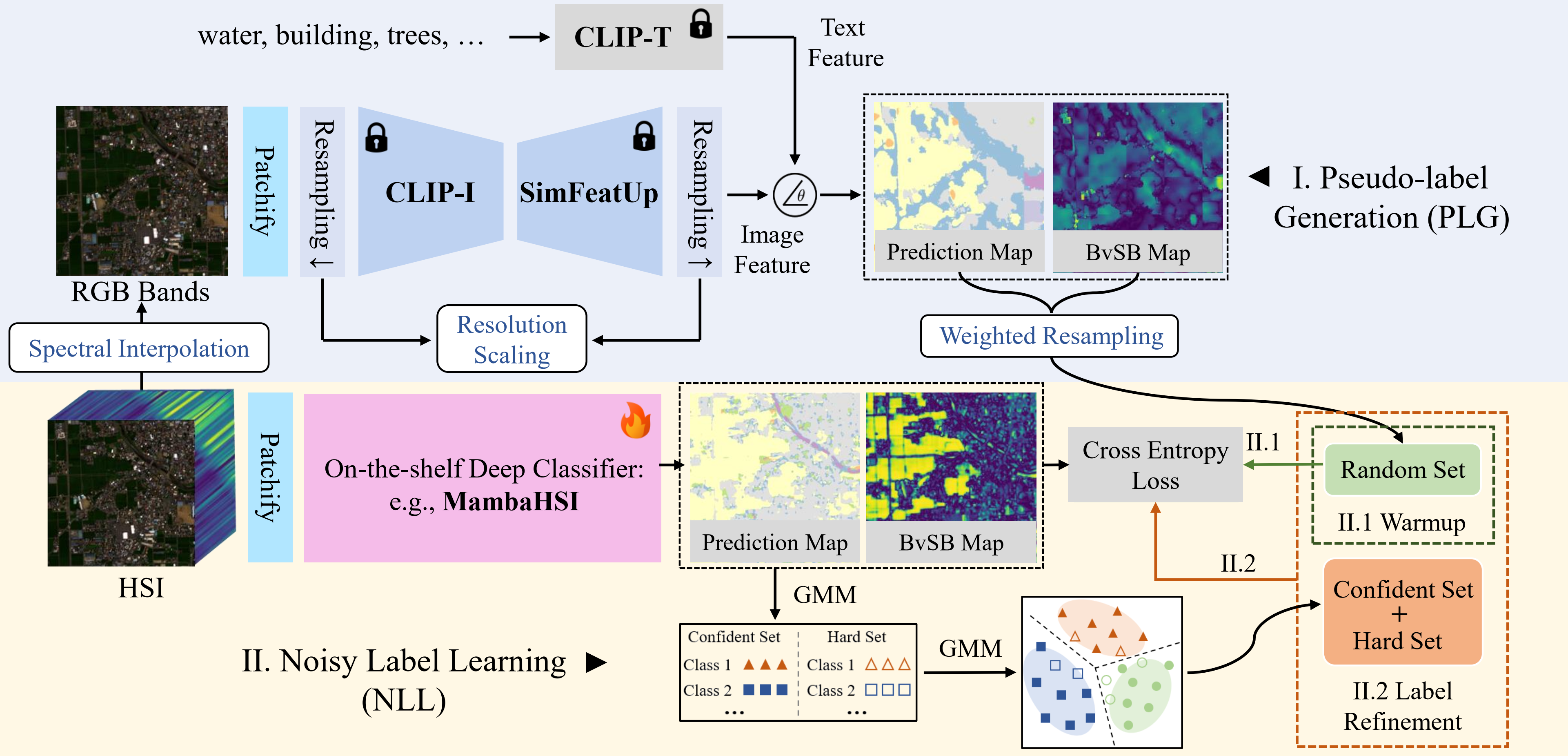}
  \caption{Overall framework of our proposed SPECIAL. The proposed framework consists of two stages: CLIP-based pseudo-labels generation (PLG) and noisy label learning (NLL). In the PLG stage, CLIP classifies interpolated RGB bands, generating pseudo-labels with confidence scores, while NLL further improves classification accuracy by incorporating spectral information with a label refinement strategy.}\label{fig:framework}
\end{figure*}

\subsection{CLIP-based Open Vocabulary Semantic Segmentation}
Contrastive Language-Image Pre-training (CLIP~\cite{radford2021learning}) learns to align images with their corresponding text descriptions in the feature space. CLIP employs a dual-encoder architecture, in which an image encoder and a text encoder are jointly trained with a contrastive objective so that matched image–text pairs are pulled close and mismatched pairs are pushed apart in a shared feature space. Once trained on large-scale web data, CLIP can directly perform zero-shot image classification by comparing the visual feature of an input image with text embeddings of category names expressed in natural language prompts, thereby enabling recognition over an essentially open vocabulary without task-specific fine-tuning. Motivated by this impressive zero-shot capability, a growing body of work attempts to transfer CLIP’s knowledge of visual concepts from image-level classification to dense prediction tasks, giving rise to CLIP-based open-vocabulary semantic segmentation, where each pixel (or patch) is assigned a label drawn from an unrestricted set of categories.

MaskCLIP~\cite{zhou2022extract} makes an early attempt to adapt CLIP to pixel-level segmentation by reusing the pretrained visual backbone and converting patch-level representations into dense predictions guided by text features, thus extracting pseudo labels for each spatial location without the need for additional human annotations. Building upon this idea, subsequent methods such as SCLIP~\cite{wang2025sclip}, GEM~\cite{bousselham2024grounding}, and ClearCLIP~\cite{lan2025clearclip} further explore how to more effectively exploit the internal structure of CLIP. In particular, they typically refine the attention mechanisms, redesign the interaction between visual and textual tokens, and introduce lightweight architectural modifications (e.g., decoder heads or feature fusion modules) to enhance the ability of CLIP features to capture fine-grained spatial details, leading to improved performance on dense prediction benchmarks under the open-vocabulary setting.

Recently, SegEarth-OV~\cite{li2024segearth} extends CLIP-based open-vocabulary segmentation to the field of remote sensing. It incorporates a novel feature upsampler, namely SimFeatUp, together with a global bias alleviation strategy to mitigate the sensitivity of CLIP to low-resolution, small-scale objects that are common in aerial and satellite imagery. This design effectively strengthens the representation of subtle structures such as narrow roads or small buildings, and demonstrates significant improvements in practical applications including building extraction and road detection. Additionally, DiffCLIP~\cite{zhang2024diffclip} has recently explored the use of CLIP for few-shot hyperspectral image (HSI) classification, showing that CLIP’s language-aligned visual features can be beneficial even when only a handful of labeled HSI samples are available. However, despite these advances, zero-shot HSI classification based on CLIP remains largely unexplored, and there is still a lack of dedicated frameworks that fully exploit CLIP’s open-vocabulary capabilities in the hyperspectral domain.


\section{Method}\label{sec:method}
The overall framework of SPECIAL is shown in Fig.~\ref{fig:framework}. The overall process can be divided into two stages, including CLIP-based pseudo-labels generation (PLG) and the noisy label learning (NLL) stage. In the PLG stage, we obtain the RGB bands by interpolating the HSI, and then a CLIP-based classification method, SegEarth-OV~\cite{li2024segearth}, is employed to classify the pixels of the RGB image, yielding pseudo-labels and corresponding confidence scores. To enhance the quality of pseudo-labels, we propose a resolution scaling strategy that fuses predictions from multiple scales to account for objects of varying sizes. In the NLL stage, there are two phases including a warmup phase and a label refinement phase. In the first phase, we train a hyperspectral classification network by sampling the pseudo-labels. In our work we adopt MambaHSI~\cite{10604894} as the spectral classifier since the model is able to model long-range interaction efficiently. In the second phase, we categorize the predicted samples of each class into confident and hard sets based on the best versus second best (BvSB)~\cite{cao2020hyperspectral,wang2023collaborative} distribution provided by MambaHSI. Then we construct class-specific probability distributions and calculate the probability density of each sample under each class distribution. These densities are normalized to obtain soft pseudo-labels, which are then incorporated into the training set, further improving the classification performance.
From a high-level perspective, SPECIAL can be interpreted as transferring the open-vocabulary semantic prior encoded in CLIP into the hyperspectral domain under the guidance of a spectral classifier. 
The PLG stage is responsible for harvesting category-aware supervision from an RGB proxy image, whereas the NLL stage focuses on correcting the noise in these pseudo-labels by exploiting the rich spectral signatures of HSI. 
Such a decoupled design makes the framework modular and flexible: more powerful CLIP variants or HSI backbones can be plugged in without redesigning the overall training paradigm. A more detailed description is provided in the following.

\subsection{CLIP-based Pseudo-label Generation}
\label{sed:plg}

In this stage, we first transform the original hyperspectral image (HSI) into a three-channel RGB-like image that can be directly processed by vision-language models. Specifically, we linearly interpolate the spectral bands around $655\,\mathrm{nm}$, $553\,\mathrm{nm}$, and $451\,\mathrm{nm}$, which approximately correspond to the red, green, and blue channels, respectively. This yields a false-color RGB image $\mathbf{I}\in\mathbb{R}^{H\times W\times 3}$ that preserves the main spatial structures and semantic cues of the HSI while remaining compatible with CLIP-based architectures. This conversion establishes a connection between the hyperspectral domain and the natural-image domain where CLIP~\cite{radford2021learning} is trained.

Next, we adopt the training-free open-vocabulary segmentation framework SegEarth-OV~\cite{li2024segearth} to obtain pixel-wise semantic scores and confidence maps from $\mathbf{I}$. SegEarth-OV employs a ViT-based CLIP image encoder (CLIP-I) and a CLIP text encoder (CLIP-T) to map images and category names into a shared embedding space. Importantly, SegEarth-OV does not directly use the final CLIP patch outputs for dense matching. Instead, to mitigate the mismatch caused by directly upsampling the final projected tokens, it extracts intermediate representation from the input tokens of the last Transformer block. Let the patch size be $P$ (e.g., $P=16$), so that $h=H/P$ and $w=W/P$. Denote by
\begin{equation}
\mathbf{X}^{(L)}=\big[\mathbf{x}^{(L)}_{\mathrm{cls}},\mathbf{x}^{(L)}_{1},\ldots,\mathbf{x}^{(L)}_{hw}\big]^{\top}\in\mathbb{R}^{(hw+1)\times d},
\end{equation}
the token sequence fed into the last Transformer block (block index $L$), where $\mathbf{x}^{(L)}_{\mathrm{cls}}$ is the global \texttt{[CLS]} token and $\mathbf{x}^{(L)}_{i}$ are patch tokens. SegEarth-OV then applies the original CLIP projection head $\mathrm{Proj}(\cdot)$ to the patch tokens of this intermediate layer to obtain low-resolution dense features:
\begin{equation}
\begin{aligned}
&\mathbf{F}_{\mathrm{LR}} = \mathrm{Proj}\big(\mathbf{X}^{(L)}_{\mathrm{patch}}\big) \in \mathbb{R}^{(hw)\times c}, \\
&\mathbf{X}^{(L)}_{\mathrm{patch}} = \big[\mathbf{x}^{(L)}_{1},\ldots,\mathbf{x}^{(L)}_{hw}\big]^{\top}.
\end{aligned}
\end{equation}
To obtain pixel-aligned representations, SegEarth-OV further employs SimFeatUp~\cite{li2024segearth} to upsample and refine $\mathbf{F}_{\mathrm{LR}}$ to the original spatial resolution:
\begin{equation}
\mathbf{F}_{\mathrm{HR}}=\mathrm{SimFeatUp}\big(\mathbf{F}_{\mathrm{LR}},\,\mathbf{I}\big)\in\mathbb{R}^{H\times W\times c},
\end{equation}
where SimFeatUp is a lightweight, once-trained feature upsampler guided by the high-resolution RGB image $\mathbf{I}$, and is used as a plug-and-play module during inference. Meanwhile, CLIP-T encodes each category name into a normalized text embedding $\mathbf{t}_k\in\mathbb{R}^{c}$, $k=1,\ldots,K$, forming $\mathbf{T}=[\mathbf{t}_1,\ldots,\mathbf{t}_K]^\top\in\mathbb{R}^{K\times c}$. After obtaining the pixel-wise visual feature $\mathbf{f}(p)\in\mathbb{R}^{c}$ at pixel $p$ from $\mathbf{F}_{\mathrm{HR}}$, we compute open-vocabulary classification scores via cosine similarity:
\begin{equation}
s_{p,k}=\cos\!\big(\mathbf{f}(p),\mathbf{t}_k\big)
      =\frac{\mathbf{f}(p)^\top \mathbf{t}_k}{\|\mathbf{f}(p)\|_2\,\|\mathbf{t}_k\|_2}.
\end{equation}
Following SegEarth-OV, we suppress the CLIP ``global bias'' by subtracting a scaled global embedding from each pixel feature:
\begin{equation}
\widetilde{\mathbf{f}}(p)=\mathbf{f}(p)-\lambda\,\mathbf{o}_{\mathrm{cls}},
\end{equation}
where $\mathbf{o}_{\mathrm{cls}}$ is the projected \texttt{[CLS]} embedding and $\lambda$ is a fixed scaling factor. Finally, we normalize the scores over all categories (with temperature $\tau$) to obtain pixel-wise class probabilities:
\begin{equation}
p_{p,k}=\frac{\exp\big(s_{p,k}/\tau\big)}{\sum_{k'=1}^{K}\exp\big(s_{p,k'}/\tau\big)}.
\end{equation}
In this way, we obtain a probability distribution for every pixel, which can be directly used for pseudo-label generation and subsequent refinement.

In order to obtain not only a hard pseudo-label (i.e., the most likely category) but also a measure of its reliability, we introduce a confidence measure based on the best versus second best (BvSB) criterion~\cite{cao2020hyperspectral,wang2023collaborative}. Formally, let
\[
p_i = (p_{i1}, p_{i2}, \ldots, p_{iC})^{\mathrm{T}}
\]
denote the classification probability vector of the $i$-th pixel $x_i$, where $C$ denotes the total number of categories and $p_{ij}$ represents the probability that pixel $x_i$ belongs to class $j$. We then define the BvSB-based confidence score for pixel $x_i$ as
\begin{equation}\label{eq:bvsb}
\text{BvSB}(x_i) = P_B(p_i) - P_{SB}(p_i),
\end{equation}
where $P_B(p_i)$ denotes the highest probability value in $p_i$ (corresponding to the predicted class), and $P_{SB}(p_i)$ denotes the second highest probability value in $p_i$. By definition, the BvSB score is non-negative and reflects the margin between the most likely and the second most likely classes. A larger BvSB value indicates that the model clearly prefers one class over all others, which means that the corresponding prediction is more reliable and less ambiguous. In contrast, when the difference between the top two probabilities is small, the model is uncertain: the pixel may lie near a decision boundary, correspond to a mixed region, or simply be misclassified. Therefore, a low BvSB value suggests that the pseudo-label assigned to this pixel should be treated carefully. Based on these confidence scores, we divide the pseudo-labeled pixels into several confidence levels by setting appropriate thresholds on the BvSB values. Pixels with high BvSB are grouped into confident subsets, while those with medium or low BvSB form hard subsets, respectively. This confidence-aware partition is important in the subsequent noise-robust learning (NLL) stage. It provides a clear way to construct clean, confident, and hard sets from the pseudo-labels. Clean and highly confident pixels can be used with strong supervision signals, which helps stabilize the optimization process, whereas hard or low-confidence pixels are used with specially designed supervisory strategies. In this manner, our framework uses CLIP-based pseudo-label generation not only to obtain semantic annotations but also to estimate their reliability, which supports more robust training with noisy pseudo-labels.

\begin{figure}[t]
  \centering
  \includegraphics[width=0.9\linewidth]{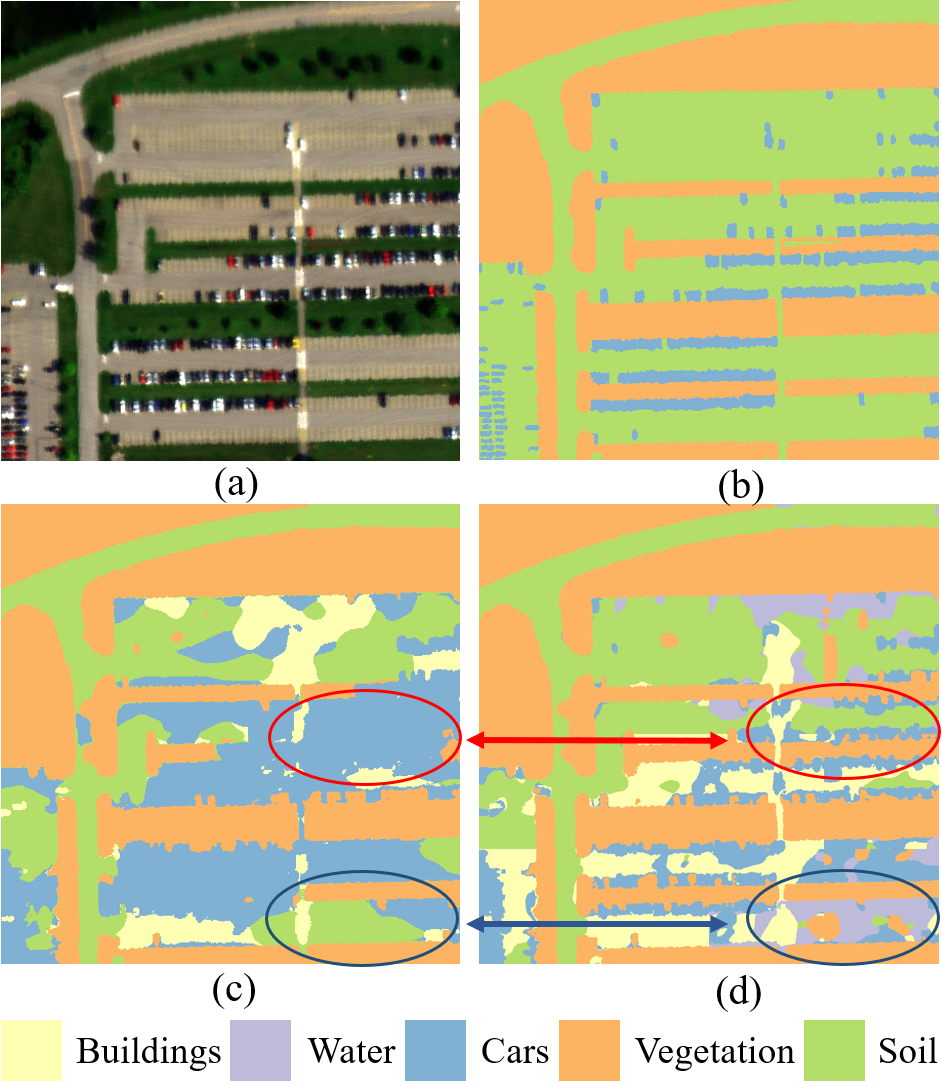}
  \caption{Comparison of prediction results at different image scales. (a) The original image. (b) Ground truth. (c) Prediction result without any image upsampling, where the model directly operates at the original resolution. (d) Prediction result with $2\times$ image upsampling before inference. The comparison shows that higher input resolution improves the detection of small objects such as individual cars and thin structures, while a lower resolution is more suitable for capturing large homogeneous areas such as continuous road regions. This indicates that different image scales emphasize complementary aspects of the scene.}
  \label{fig:scale}
\end{figure}

However, unlike images in natural scenes, remotely sensed HSIs often have very large spatial dimensions, sometimes covering entire cities or regions with very high resolutions. Directly feeding such large images into CLIP-based models is computationally expensive and requires a large amount of memory. A commonly used strategy to reduce this cost is the \emph{patchify} operation, in which the large-scale imagery is divided into a set of smaller, possibly overlapping patches using a sliding window. Each patch is processed independently by the model to obtain a local prediction map, and the final prediction for the whole image is reconstructed by combining the predictions of all patches.

Although this patch-based strategy is practical, it has several limitations in the context of remote sensing segmentation. Because the receptive field of CLIP is fixed and object scales in remote sensing scenes vary widely, the model may fail to capture both global context and fine local details at the same time. If the patch size is small, large-scale structures (e.g., long roads or large parking lots) may be fragmented or only partially visible within each patch, which leads to incomplete or inconsistent predictions. On the other hand, if the patch size is large, small objects (e.g., cars, narrow alleys, or thin roads) may occupy only a small region of the patch and thus be dominated by the surrounding background, causing them to be missed or misclassified. In addition, patch-wise processing can introduce artifacts at patch boundaries, which further harms the spatial consistency of the final prediction map.

The influence of input scale on CLIP-based segmentation is illustrated in Fig.~\ref{fig:scale}. When the image resolution is relatively low, CLIP models tend to capture the global layout and identify large semantic regions such as parking lots and continuous roads, but they may miss local details like thin or discontinuous road segments. In contrast, when the image is upsampled to a higher resolution, the model can more easily recognize small objects such as cars and other fine structures, but large objects such as long roads may not appear at a suitable scale, which may cause them to be fragmented or incorrectly classified. This trade-off shows that a single resolution is not sufficient to achieve good performance on both small and large structures in complex remote sensing scenes.

To address these problems and better use the complementary information at different image scales, we propose a simple but effective resolution scaling (RS) strategy that fuses prediction results obtained at multiple resolutions. Given an input image $\mathbf{I}$, we first generate several scaled versions of $\mathbf{I}$ by applying different upsampling factors $\{s_1, s_2, \ldots, s_N\}$. For each scale $s_i$, we upsample the image using an upsampling operator $\mathbf{U}_{s_i}$, feed the upsampled image into the CLIP-based segmentation pipeline to obtain a probability map at that scale, and then map the probability map back to the original resolution using a downsampling operator $\mathbf{D}_{s_i}$. The final prediction $\mathbf{P}$ is obtained by averaging the probability maps across all scales:
\begin{equation}\label{eq:scale}
\mathbf{P} = \frac{1}{N}\sum_{i=1}^{N} \mathbf{D}_{s_{i}}(\text{CLIP}(\mathbf{U}_{s_{i}}(\mathbf{I}))), 
\end{equation}
where $s_i$ denotes the upsampling factor at the $i$-th scale, and $\mathbf{D}$ and $\mathbf{U}$ represent the downsampling and upsampling operations, respectively.

In our work, both upsampling and downsampling operations are implemented using bicubic interpolation, which offers a good balance between computational efficiency and interpolation quality. Bicubic interpolation is widely used in image processing and usually introduces few artifacts, which makes it suitable for our multi-scale fusion strategy. In principle, more advanced single-image super-resolution methods or learned upsampling modules could replace the bicubic interpolation operators $\mathbf{U}_{s_i}$ and $\mathbf{D}_{s_i}$ and may further improve performance, especially for very small objects. However, our experiments show that even this simple choice of interpolation already brings clear improvements.

\subsection{Noisy Label Learning}
In this stage, we explicitly incorporate spectral information to further improve the classification accuracy obtained from the previous pseudo-label generation stage. The overall training process consists of two phases: a \emph{warmup phase} and a \emph{label refinement phase}. Compared with the RGB proxy used in the PLG stage, the original hyperspectral measurements contain rich material-specific spectral signatures that are more stable across changes in illumination, season, and atmospheric conditions. By directly modelling these spectral distributions, the noisy label learning (NLL) stage can correct systematic biases inherited from CLIP-based pseudo-labels and adapt the decision boundaries to the statistical properties of the target scene.

More concretely, the pseudo-labels produced by SegEarth-OV are generated in the RGB space and inevitably suffer from domain gaps between natural images and remote sensing data. These domain gaps lead to label noise, especially for confusing categories or boundary regions. In contrast, hyperspectral pixels from different classes often exhibit distinct spectral curves, which provide complementary information that is not fully exploited in the PLG stage. Therefore, our NLL stage aims to transfer the coarse semantic prior from the RGB-based pseudo-labels into a purely spectral model, and then refine these labels by exploiting the underlying spectral distributions.

\subsubsection{Warmup phase}
In the first warmup training phase, we use the pseudo-labels provided by SegEarth-OV to train an HSI classification network, namely MambaHSI~\cite{10604894}, with the standard cross-entropy loss. MambaHSI is a recent hyperspectral classification model that directly operates on spectral vectors and can capture long-range dependencies along the spectral dimension. Since the input of MambaHSI is the original hyperspectral data rather than the interpolated RGB image, this warmup training encourages the model to learn a mapping from spectral signatures to the semantic categories implied by CLIP-based pseudo-labels.

Due to the extreme class imbalance and significant label noise present in remote sensing scenes, directly training on all pseudo-labeled samples may cause the model to be dominated by large classes and overfit to noisy labels. To avoid this, we adopt a class-balanced sampling strategy: in each training iteration, a fixed number of samples is randomly drawn from each class, instead of using all available labeled pixels. The resulting mini-batch, together with their pseudo-labels, is referred to as the \emph{random set}. This strategy ensures that each class is observed with a similar frequency during training, increases the diversity of the training samples, and prevents the model from relying too heavily on a few dominant categories.

Furthermore, we make use of the BvSB values provided by SegEarth-OV as sampling weights. Samples with higher BvSB scores, which correspond to more confident predictions, are more likely to be selected, while those with low confidence are sampled less frequently. Each class is sampled independently, so that both rare classes and common classes benefit from a focus on reliable pseudo-labels. Overall, this warmup process can be regarded as distilling the coarse semantic prior from the PLG stage into a spectral model that no longer depends on the RGB proxy. 

\subsubsection{Label refinement phase}
After the warmup phase, the spectral classifier already captures useful information from the noisy pseudo-labels. However, the remaining label noise still limits the final performance. To further reduce the influence of noisy labels and exploit the spectral structure of the data, we introduce a label refinement phase. In this phase, we construct two additional training sets, namely the \emph{confident set} and the \emph{hard set}, and assign soft labels to their samples based on a class-wise Gaussian Mixture Model (GMM). The overall procedure consists of the following four steps:

\begin{enumerate}
    \item For the region of interest, we first use the warmup-trained MambaHSI network to obtain predicted labels and the corresponding BvSB values for all pixels. For each class, we collect the predicted samples belonging to that class and apply a Gaussian Mixture Model (GMM) to the distribution of their BvSB values. The GMM typically contains two components, which naturally divide the samples into a high-confidence group and a low-confidence group. The spectral samples associated with the high-confidence group are termed the \emph{confident set}, while those associated with the low-confidence group form the \emph{hard set}. The confident set is expected to be relatively clean, containing mostly correctly labeled pixels, whereas the hard set is more informative but may contain a higher proportion of mislabeled samples. This step separates easy and hard examples in an adaptive way rather than relying on a fixed confidence threshold.
    
    \item Next, we perform principal component analysis (PCA) on the spectral vectors to reduce their dimensionality. For each class, we then use a GMM to fit the distribution of the reduced spectral features from the confident set of that class. In this way, we obtain a class-conditional density model that describes how clean spectral samples of that class are distributed in the low-dimensional PCA space.
    
    \item Based on the fitted GMMs, we further assign soft labels to both confident and hard samples. For each sample $x$ (represented by its reduced spectral feature), we compute its probability density under the distribution of each class. These densities are then normalized across all classes to form a probability vector, which is used as a soft label. In other words, instead of assigning a single hard class label, we estimate how likely the sample belongs to each class according to the learned spectral distributions.
    
    \item Finally, we include samples from the random set, confident set, and hard set into the training process. The random set preserves the semantic prior from the PLG stage, the confident set provides relatively clean supervision, and the hard set encourages the model to explore ambiguous regions and refine its decision boundaries. The generated soft labels for the confident and hard sets help to down-weight unreliable classes and provide a tradeoff between exploration (learning from difficult samples) and exploitation (trusting reliable spectral evidence).
\end{enumerate}

Mathematically, the soft label $\tilde{y}(x)$ for a sample with reduced spectral feature $x$ is calculated as
\begin{equation}\label{eq:sl}
\tilde{y}_k(x) = \frac{\sum_{m=1}^{M} \pi_{km} \mathcal{N}(x \mid \mu_{km}, \Sigma_{km})}{\sum_{k'=1}^{K} \sum_{m'=1}^{M} \pi_{k'm'} \mathcal{N}(x \mid \mu_{k'm'}, \Sigma_{k'm'})},
\end{equation}
where $K$ is the number of classes, $M$ is the number of Gaussian components per class, and $\pi_{km}$, $\mu_{km}$, and $\Sigma_{km}$ represent the weight, mean, and covariance of the $m$-th Gaussian component of the $k$-th class, respectively. The term $\mathcal{N}(x \mid \mu_{km}, \Sigma_{km})$ denotes the value of the Gaussian density function at $x$. The $k$-th component $\tilde{y}_k(x)$ is thus the normalized class-conditional density and can be interpreted as the probability that $x$ belongs to class $k$ under the GMM model. This construction naturally produces softer labels for samples located in overlapping regions and sharper labels for samples lying in high-density regions of a particular class.

The total training loss of the NLL stage is defined as
\begin{equation}\label{eq:loss}
L = \text{CE}(\hat{y}_r, \tilde{y}_r) + \lambda_1 \text{CE}(\hat{y}_c, \tilde{y}_c) + \lambda_2 \text{CE}(\hat{y}_h, \tilde{y}_h),
\end{equation}
where $\text{CE}$ denotes the cross-entropy loss, $\hat{y}_r$, $\hat{y}_c$, and $\hat{y}_h$ are the predicted class probability vectors for the random, confident, and hard sets, respectively, $\tilde{y}_r$ is the pseudo-label provided by SegEarth-OV for the random set, and $\tilde{y}_c$ and $\tilde{y}_h$ are the soft labels generated by the GMM for the confident and hard sets. The scalars $\lambda_1$ and $\lambda_2$ are hyperparameters that control the relative contributions of the confident and hard sets to the overall objective.
In this loss, the first term encourages the model to be consistent with the original CLIP-based pseudo-labels and maintains the semantic prior learned from the PLG stage. The second term exploits the relatively clean confident set with soft labels, which helps to refine the decision boundaries using reliable spectral evidence. The third term allows the model to benefit from the informative but noisy hard set, while the soft labels and weighting factor $\lambda_2$ reduce the risk of overfitting to incorrect labels. Together, these three terms enable the model to gradually correct noisy labels, sharpen the class boundaries in the spectral space, and achieve robust classification performance under severe label noise.

 \begin{table}[t]
\centering
\caption{The number of the labeled pixels in different datasets.}
\label{tab:data}
\renewcommand\arraystretch{1.2}
\small
\resizebox{1\linewidth}{!}{
\begin{tabular}{cc|cc|cc}
\Xhline{1pt}
\multicolumn{2}{c|}{Chikusei}                & \multicolumn{2}{c|}{Pavia Centre} & \multicolumn{2}{c}{AeroRIT}                 \\ \hline
Category             & Total                 & Category                & Total   & Category             & Total                \\ \hline
Farmland             & 16289                 & Water                   & 65971   & Buildings            & 536188               \\
Soil                 & 7458                  & Trees                   & 7598    & Water                & 46480                \\
Road                 & 801                   & Meadows                 & 3090    & Cars                 & 91268                \\
Grass                & 11852                 & Self-Blocking Bricks    & 2685    & Vegetation           & 1474345              \\
Trees                & 20516                 & Bare Soil               & 6584    & Soil                 & 993792               \\
Buildings            & 3604                  & Asphalt                 & 9248    &                      &                      \\
Water                & 2586                  & Bitumen                 & 7287    &                      &                      \\
\multicolumn{1}{l}{} & \multicolumn{1}{l|}{} & Tile                    & 42826   & \multicolumn{1}{l}{} & \multicolumn{1}{l}{} \\
\multicolumn{1}{l}{} & \multicolumn{1}{l|}{} & Shadows                 & 2863    & \multicolumn{1}{l}{} & \multicolumn{1}{l}{} \\ \Xhline{1pt}
\end{tabular}
}
\end{table}

\begin{figure*}[t]
  \centering
  \includegraphics[width=0.9\textwidth]{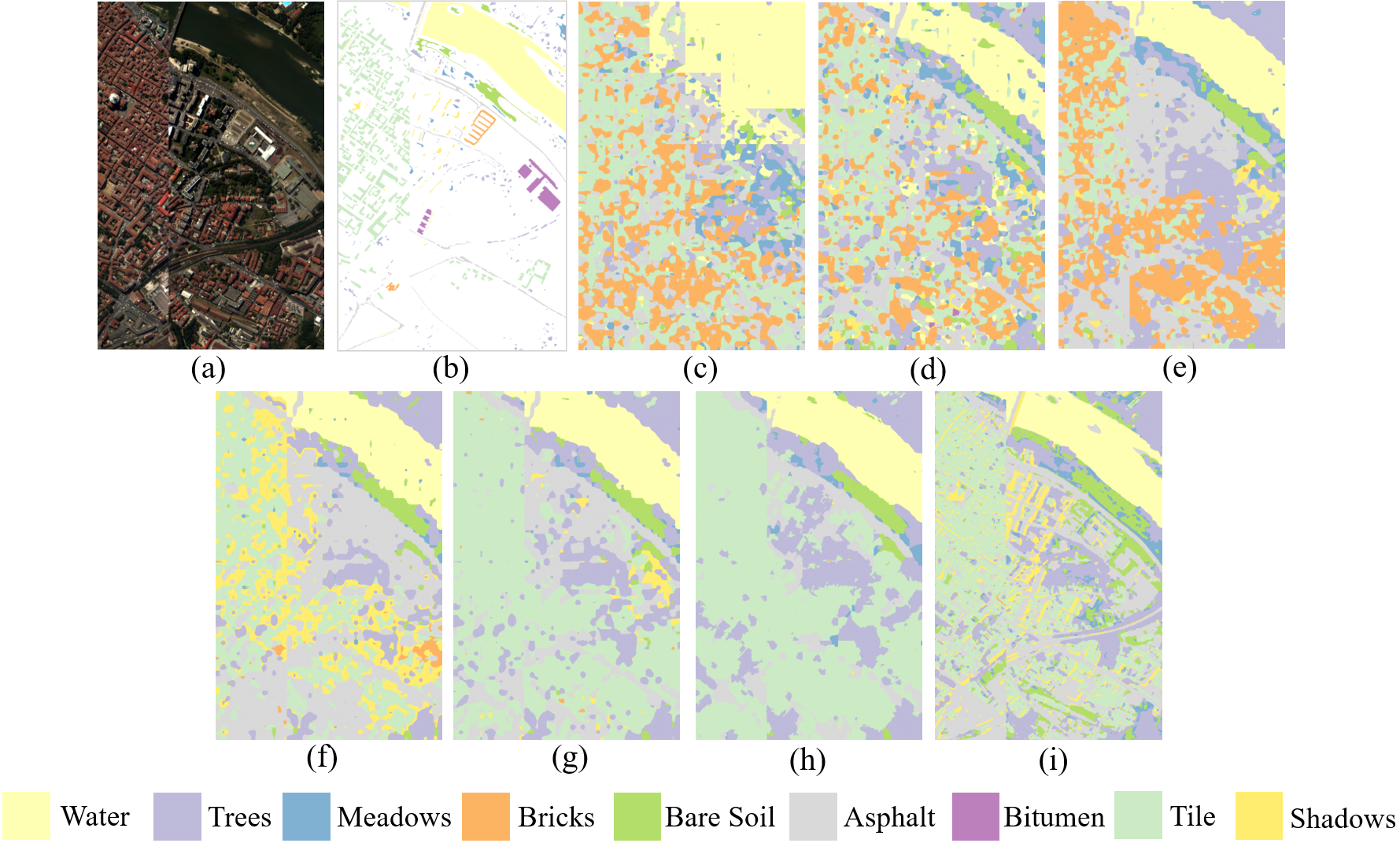}
  \caption{Visualization of the classification maps produced by different approaches on the Pavia Centre dataset. (a) False-color image derived from the hyperspectral data. (b) Ground truth reference map. (c) CLIP. (d) MaskCLIP. (e) SCLIP. (f) GEM. (g) ClearCLIP. (h) SegEarth-OV. (i) The proposed method (Ours). The comparison shows that our method generates cleaner and more coherent segmentation maps, preserves object boundaries better, and reduces misclassifications in complex urban regions compared with existing CLIP-based or open-vocabulary baselines.}
  \label{fig:paviac}
\end{figure*}

\begin{table*}[t]
\centering
\caption{Classification results of different methods on the Pavia Centre dataset. The best class-specific, OA, AA, and $\kappa$ values are in \textbf{bold}.}
\label{tab:paviac}
\renewcommand\arraystretch{1.1}
\small
\resizebox{1\textwidth}{!}{
\begin{tabular}{c|ccccccccc|ccc}
\Xhline{1pt}
Method   & Water  & Trees & Meadows & Bricks & Bare Soil & Asphalt & Bitumen & Tile  & Shadows & OA(\%) & AA(\%) & $\kappa$(\%) \\ \hline
CLIP~\cite{radford2021learning}      & 83.73  & 1.42  & 0.32    & \textbf{12.22}  & 0.00      & 8.46    & 0.00    & 53.86 & 0.45    & 53.69 & 17.83 & 35.29 \\
MaskCLIP~\cite{zhou2022extract}  & 91.71  & 83.25 & \textbf{77.38}   & 1.45   & 66.40     & 85.22   & 0.00    & 60.48 & 4.51    & 72.59 & 52.27 & 62.90 \\
SCLIP~\cite{wang2025sclip}     & 99.45  & 85.25 & 16.70   & 0.00   & 43.26     & 88.47   & 0.00    & 45.23 & 1.15    & 69.55 & 42.17 & 59.02 \\
GEM~\cite{bousselham2024grounding}       & 98.35  & 82.80 & 24.17   & 0.00   & 68.92     & 96.84   & 0.00    & 75.32 & 27.17    & 79.95 & 52.62 & 72.21 \\
ClearCLIP~\cite{lan2025clearclip} & \textbf{99.97}  & 91.60 & 31.65   & 0.00   & 65.01     & 89.10   & 0.00    & 99.04 & 3.11    & 87.02 & 53.28 & 81.33 \\
SegEarth-OV~\cite{li2024segearth}  & 99.93  & 92.98 & 13.46   & 0.00   & 76.17     & 63.47   & 0.00    & \textbf{99.42} & 0.14    & 85.64 & 49.51 & 79.00 \\
Ours      & 98.70  & \textbf{98.99} & 34.30   & 0.00   & \textbf{98.31}     & \textbf{99.46}   & 0.00    & 98.24 & \textbf{98.71}    & \textbf{90.62} & \textbf{69.63} & \textbf{86.75} \\ \Xhline{1pt}
\end{tabular}
}
\end{table*}

\begin{figure*}[h]
  \centering
 \includegraphics[width=0.9\textwidth]{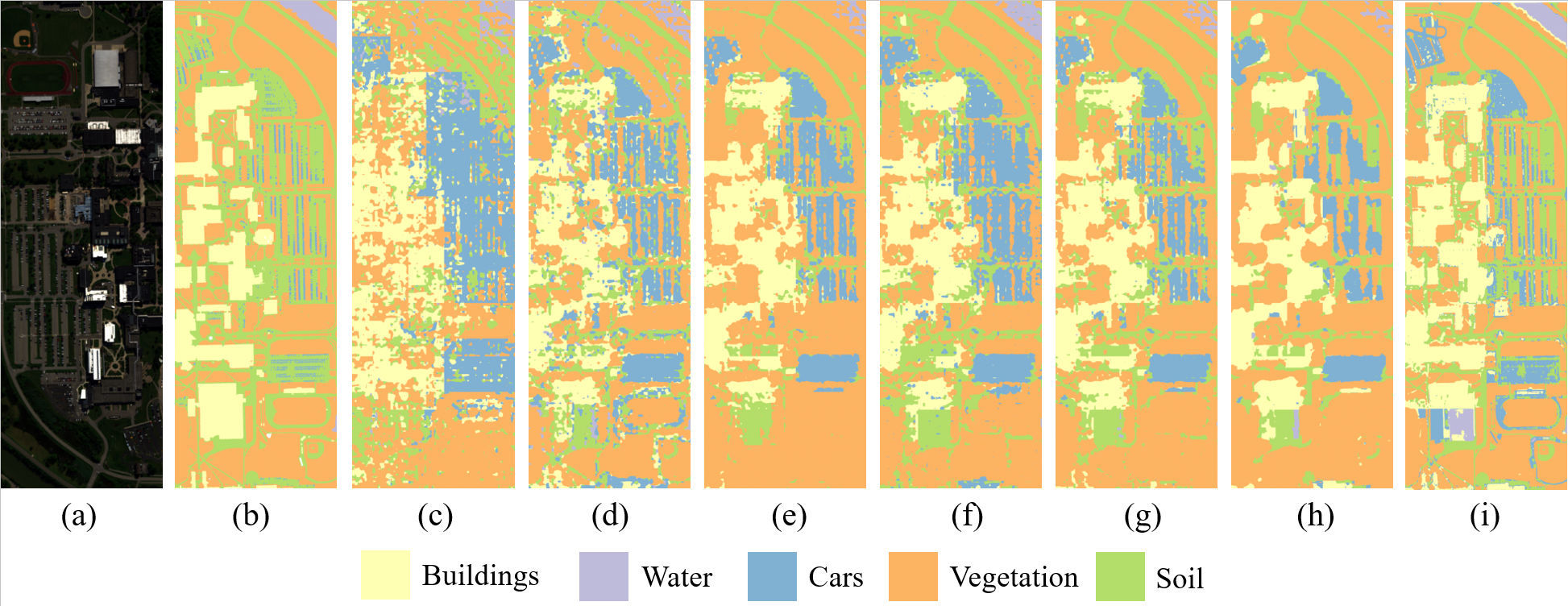}
    \caption{Visualization of the classification maps provided by different approaches on the AeroRIT dataset. (a) False color image. (b) Ground truth. (c) CLIP. (d) MaskCLIP.	(e) SCLIP. (f) GEM.	(g) ClearCLIP. (h) SegEarth-OV. (i) Ours.}\label{fig:aerorit}
\end{figure*}

\begin{table*}[t]
\centering
\caption{Classification results of different methods on the AeroRIT dataset. The best class-specific, OA, AA, and $\kappa$ values are in \textbf{bold}.}
\label{tab:aerorit}
\renewcommand\arraystretch{1.1}
\small
\resizebox{0.75\textwidth}{!}{
\begin{tabular}{c|ccccc|ccc}
\Xhline{1pt}
Method   & Buildings & Water & Cars  & Vegetation & Road  & OA(\%) & AA(\%) & $\kappa$(\%) \\ \hline
CLIP~\cite{radford2021learning}      & 34.72     & 31.99 & 44.92 & 34.78      & 6.18  & 25.97  & 30.52  & -1.29        \\
MaskCLIP~\cite{zhou2022extract}  & 69.05     & 51.42 & 92.37 & 80.75      & 34.32 & 63.97  & 65.58  & 48.74        \\
SCLIP~\cite{wang2025sclip}     & 67.62     & 26.07 & 88.38 & 93.35      & 20.74 & 64.86  & 59.23  & 46.27        \\
GEM~\cite{bousselham2024grounding}        & 68.72     & 68.18 & \textbf{92.91} & 82.90      & 37.37 & 66.15  & 70.02  & 51.13        \\
ClearCLIP~\cite{lan2025clearclip} & 72.14     & 33.28 & 90.90 & 88.96      & 38.83 & 69.47  & 64.82  & 54.28        \\
SegEarth-OV~\cite{li2024segearth}  & 75.60     & 20.35 & 70.84 & 94.81      & 33.48 & 70.34  & 59.02  & 54.48        \\
Ours      & \textbf{84.57} & \textbf{79.45} & 78.69 & \textbf{95.28} & \textbf{61.91} & \textbf{82.18} & \textbf{79.98} & \textbf{73.83} \\ \Xhline{1pt}
\end{tabular}
}
\end{table*}

\section{Experimental Results}
\label{sec:exp}
\subsection{Dataset}
We conducted experiments on three publicly available datasets, including Chikusei~\cite{NYokoya2016}, AeroRIT~\cite{rangnekar2020aerorit}, and Pavia Centre~\cite{plaza2009recent}. A brief introduction is presented as follows.

1) \textbf{Pavia Centre}: The Pavia Centre dataset was acquired by the ROSIS sensor during a flight campaign over Pavia, northern Italy. It contains 102 spectral bands covering the spectral range of 430–860~nm, with a size of $1095\times 751$. The geometric resolution is 1.3 meters. There are nine land cover categories in the dataset, including water, trees, meadows, bricks, bare soil, asphalt, bitumen, tile, and shadows.

2) \textbf{AeroRIT}: The AeroRIT dataset was captured with a visible near-infrared (VNIR) hyperspectral Headwall Photonics Micro Hyperspec E-Series CMOS sensor. It contains 372 spectral bands covering the spectral range of 397–1003~nm. The ground sampling distance is about 0.4 meters. We use the center area with $1024\times 3072$ pixels for our experiment to exclude invalid areas. In addition, we removed the first and last 10 spectral bands due to significant data noise in these bands. There are five distinct land cover categories in the dataset, including buildings, water, cars, vegetation, and road.

3) \textbf{Chikusei}: The dataset was captured using a Headwall Nano-Hyperspec sensor. It contains 128 spectral bands covering the spectral range of 343–1018~nm, and the ground sampling distance is 2.5 meters. We consolidated highly similar classes within the dataset labels, such as "Bare soil (farmland)" and "Bare soil (park)", resulting in 7 distinct land cover classes, including farmland, bare soil, water, road, grass, trees, and buildings. We use the center area with $2048\times 2048$ pixels for the experiment to exclude invalid areas.

Table~\ref{tab:data} summarizes the number of labeled pixels for each class in the three datasets. It can be observed that there exists a significant class imbalance, which poses challenges for model training.
Given the considerable impact of prompt engineering on CLIP-based pseudo-labeling~\cite{zhang2024diffclip}, we deliberately designed prompts that are semantically refined rather than strictly aligned with the category names. Specifically, when generating pseudo-labels using CLIP, we replaced “Water” with “River or lake” as the prompt. This is because CLIP was trained on natural images and using “Water” directly could cause confusion, often misclassifying blue-colored buildings as water bodies. Additionally, we replaced “Vegetation” with “Grass or Trees”, which provides a more specific and meaningful description, improving the quality of the generated pseudo-labels.

\subsection{Implementation Details And Evaluation Metrics}
The hyperspectral classifier (i.e., MambaHSI) was trained for 10, 15, and 20 epochs on AeroRIT, Chikusei, and Pavia Centre, respectively, with 20 iterations per epoch. The initial learning rate was set to 1e-3 for AeroRIT, 2e-3 for Chikusei, and 4e-4 for Pavia Centre, and decayed using cosine annealing, with the minimum learning rate $\eta_{\min}$ set to 1e-5 for Chikusei and AeroRIT, and 1e-4 for Pavia Centre. In the first half of training, the model was trained with pseudo-labels generated by CLIP (i.e., SegEarth-OV). After half of the training iterations, the confident set and hard set were incorporated for training.  We adopted Adam as the optimizer. When generating pseudo-labels using SegEarth-OV, we adopted a window size of $224\times 224$ and a stride of 112 for the Chikusei and Pavia Centre datasets, and a window size of $448\times 448$ with a stride of 224 for the AeroRIT dataset due to higher ground resolution. For the resolution scaling proposed in Section~\ref{sed:plg}, we fused the prediction results from $1\times$ and $2\times$ upsampled images. To enhance the generalization ability of the network and prevent overfitting, we added Gaussian noise with a standard deviation of 0.1 to the HSI input during training. Three commonly used evaluation metrics, including overall accuracy (OA), average accuracy (AA), and kappa coefficient ($\kappa$) are adopted to evaluate the classification performance.

\subsection{Comparison With SOTAs}
\begin{figure*}[t]
  \centering
  \includegraphics[width=0.9\textwidth]{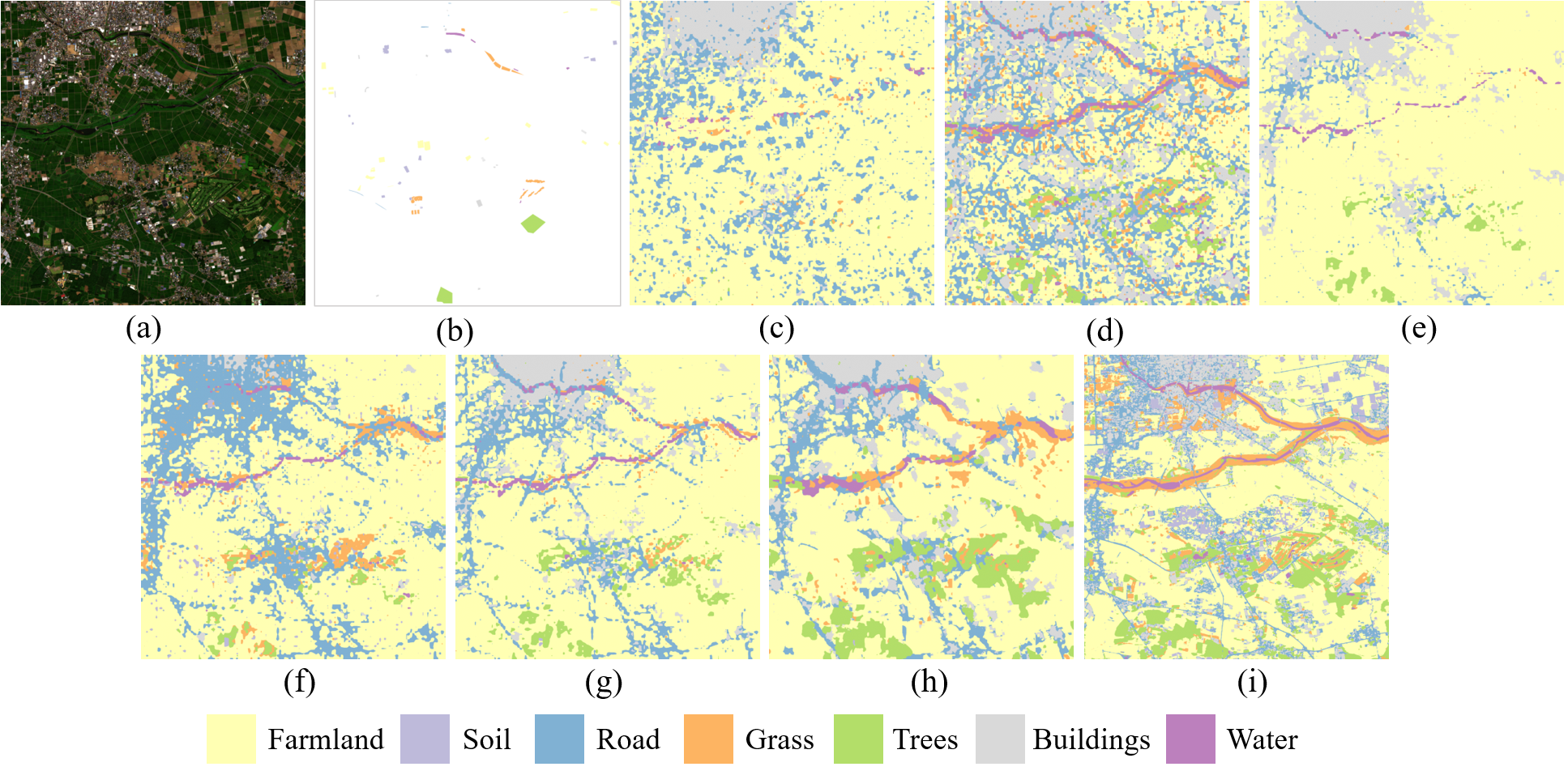}
  \caption{Visualization of the classification maps provided by different approaches on the Chikusei dataset. (a) False color image. (b) Ground truth. (c) CLIP. (d) MaskCLIP.	(e) SCLIP. (f) GEM.	(g) ClearCLIP. (h) SegEarth-OV. (i) Ours.}\label{fig:chikusei}
\end{figure*}

We compare our method with several state-of-the-art CLIP-based classification methods, including CLIP~\cite{radford2021learning}, MaskCLIP~\cite{zhou2022extract}, SCLIP~\cite{wang2025sclip}, GEM~\cite{bousselham2024grounding}, ClearCLIP~\cite{lan2025clearclip}, and SegEarth-OV~\cite{li2024segearth}. For a fair comparison, all approaches are evaluated under the same zero-shot setting and on the same three benchmark datasets, namely Pavia Centre, AeroRIT, and Chikusei, using identical class definitions and prompts. The quantitative results are summarized in Table~\ref{tab:paviac}--\ref{tab:chikusei}, while representative qualitative results are shown in Fig.~\ref{fig:paviac}--\ref{fig:chikusei}. These visualizations help to intuitively assess the spatial coherence of the predicted maps and the capability of each method in capturing object boundaries and small structures.

As can be seen from the tables, our method consistently outperforms all competing approaches in terms of overall accuracy (OA), average accuracy (AA), and kappa coefficient ($\kappa$) on all three datasets. In particular, on the Pavia Centre dataset, our framework achieves a noticeable improvement in OA and $\kappa$ compared with the best CLIP-based baseline, indicating that the proposed approach yields more reliable pixel-wise predictions over the entire scene. On the AeroRIT dataset, which contains complex urban structures and small objects such as cars, our method exhibits a larger performance gain, reflecting its improved ability to handle fine-scale details and strong inter-class confusion. A similar trend can be observed on the Chikusei dataset, where our model achieves the highest OA, AA, and $\kappa$, demonstrating that it generalizes well across different sensors, spatial resolutions, and scene types.

From the qualitative comparisons in Fig.~\ref{fig:paviac}--\ref{fig:chikusei}, it is clear that our method produces smoother and more coherent classification maps, with fewer isolated misclassified pixels and clearer region boundaries. Competing CLIP-based methods often suffer from noisy predictions and fragmented regions, especially in areas where the CLIP features at low resolution are not sufficient to distinguish visually similar classes. In contrast, our results show more homogeneous regions for large land-cover types (e.g., meadows, farmland, vegetation) and better preservation of narrow or small structures such as roads and cars. These observations are consistent with the quantitative improvements in AA, which measures the average performance across classes and is sensitive to the accuracy on rare or difficult categories.

\begin{table*}[h]
\centering
\caption{Classification results of different methods on the Chikusei dataset. The best class-specific, OA, AA, and $\kappa$ values are in \textbf{bold}.}
\label{tab:chikusei}
\renewcommand\arraystretch{1.1}
\small
\resizebox{0.8\textwidth}{!}{
\begin{tabular}{c|ccccccc|ccc}
\Xhline{1pt}
Method   & Farmland & Soil  & Road  & Grass & Trees  & Buildings & Water & OA(\%) & AA(\%) & $\kappa$(\%) \\ \hline
CLIP~\cite{radford2021learning}      & 60.06    & 0.00  & 0.00  & 0.00  & 0.00   & 2.72      & 0.00  & 15.66    & 8.97     & -4.84          \\
MaskCLIP~\cite{zhou2022extract}  & 66.95    & 18.13 & 66.54 & 71.46 & 86.68  & 17.65     & 78.58 & 66.10    & 58.00    & 56.50          \\
SCLIP~\cite{wang2025sclip}     & \textbf{96.80} & 0.00  & 7.87  & 3.58  & 77.19  & 17.59     & 60.75 & 54.35    & 37.68    & 38.49          \\
GEM~\cite{bousselham2024grounding}       & 92.79    & 20.07 & 70.41 & 40.15 & 57.99  & 8.44      & 62.95 & 56.67    & 50.40    & 43.78          \\
ClearCLIP~\cite{lan2025clearclip} & 94.92    & 10.58 & 37.58 & 27.23 & 83.24  & 12.60     & 72.39 & 62.09    & 48.36    & 49.54          \\
SegEarth-OV~\cite{li2024segearth}  & 87.34    & 26.66 & 43.70 & 66.44 & 99.76  & 28.08     & 67.75 & 75.54    & 59.96    & 67.49          \\
Ours      & 74.52    & \textbf{98.85} & \textbf{84.64} & \textbf{90.78} & \textbf{99.99} & \textbf{47.00} & \textbf{100.00} & \textbf{88.33} & \textbf{85.11} & \textbf{84.98} \\ \Xhline{1pt}
\end{tabular}
}
\end{table*}

The performance improvement can mainly be attributed to three aspects of our framework: (1) the incorporation of spectral information in the NLL stage, (2) the resolution scaling (RS) strategy in the PLG stage, and (3) the label refinement strategy based on Gaussian Mixture Models. First, instead of relying solely on RGB proxies, our method explicitly uses full hyperspectral measurements in the noisy label learning stage. Hyperspectral pixels from the same class often share similar spectral signatures, and this spectral homogeneity naturally reduces the impact of mislabeled samples. Even when the pseudo-labels from CLIP contain noticeable errors, the spectral classifier can still learn robust decision boundaries by exploiting the relationships between spectrally close samples. Second, the RS strategy fuses predictions from multiple resolutions, which helps to balance global context and local detail: low-resolution inputs emphasize large-scale structures and smooth region shapes, while high-resolution inputs highlight small objects and thin structures. By averaging predictions from different scales, we obtain more stable and accurate pseudo-labels, which provide a better starting point for subsequent spectral learning. Third, the label refinement strategy uses class-wise GMMs and BvSB-based confidence to construct random, confident, and hard sets, and assigns soft labels to the latter two. This procedure allows the model to take advantage of informative but noisy samples while reducing the risk of overfitting to incorrect labels. A typical example that highlights the benefit of our design is the \emph{water} class on the Chikusei and AeroRIT datasets. In these scenes, CLIP-provided pseudo-labels tend to misclassify many water-body pixels, often confusing them with dark roofs, shadows, or other spectrally different but visually similar regions in the RGB representation. Our NLL stage, however, can effectively learn from a small number of highly confident water samples and then generalize to other water regions by leveraging the distinctive spectral profile of water. As a result, our method achieves a significant accuracy gain on the water class compared with all CLIP-based baselines. This improvement is clearly reflected in both the class-specific accuracies and the visual results, where water bodies are more completely and cleanly delineated. In contrast, existing CLIP-based classification methods operate directly on upsampled CLIP features and struggle to recover fine-grained pixel-level semantics from low-resolution feature maps. Since the spatial resolution of CLIP features is relatively low, information about small or thin structures is often lost during feature extraction. When these features are simply upsampled, the resulting dense predictions tend to be over-smoothed and fail to capture subtle boundaries. Moreover, these methods do not explicitly exploit hyperspectral information, and thus cannot use spectral similarities to correct misclassifications introduced by noisy pseudo-labels. This limitation explains why they show inferior performance, particularly in challenging categories and complex urban scenes.

It is also important to note that, although our model significantly improves the overall classification performance, it still struggles to accurately recognize some difficult classes, such as \emph{brick} and \emph{bitumen} in the Pavia Centre dataset. The main reasons are as follows. First, these classes occupy a small proportion of the pixels, which makes them underrepresented during training. Second, their visual and spectral characteristics are close to those of other classes such as tiles or asphalt, leading to ambiguous semantic information and strong confusion even for human observers. Third, the underlying CLIP model (SegEarth-OV), on which our pseudo-labels are based, already fails to reliably distinguish these classes at the feature level. As a consequence, the pseudo-labels for brick and bitumen are highly noisy, and this noise is difficult to correct completely even with our spectral refinement strategy. These limitations in the CLIP backbone are inevitably reflected in our final results.

Despite these remaining challenges, our method still shows a clear advantage in terms of both overall accuracy and visual quality across all tested datasets. The combination of CLIP-based semantic priors, multi-scale pseudo-label generation, and noise-robust spectral learning forms an effective framework for zero-shot hyperspectral image classification. We believe that further improvements could be achieved by designing more discriminative prompts for ambiguous classes, incorporating additional spatial-context modeling, or replacing the current CLIP backbone with stronger vision-language models. We leave these directions for future work.

\subsection{Ablation Study}

In this subsection, we conduct a series of ablation experiments to better understand the contribution of each main component in our framework. Specifically, we focus on three key factors: (1) the use of full hyperspectral information instead of RGB images, (2) the resolution scaling (RS) strategy for generating pseudo-labels, and (3) the proposed label refinement strategy based on random, confident, and hard sets with soft labels. Unless otherwise stated, all experiments follow the same settings as in the main results, including the network backbone, training schedule, and evaluation metrics. 

\subsubsection{Spectral Information Incorporation}
We first study the impact of spectral information on land-cover classification performance. To this end, we construct a variant of our method in which the hyperspectral images (HSIs) used in the noisy label learning stage are replaced by their corresponding RGB images. In other words, we only keep three bands for each pixel, and all other settings are kept unchanged. Since the input dimensionality is reduced from a large number of bands to three channels, the number of principal components in the PCA step is also set to $3$ to match the RGB format. Apart from this adjustment, the network architecture, training pipeline, pseudo-label generation, and label refinement procedures remain the same.

\begin{table}[t]
\centering
\caption{Ablation results on the spectral incorporation.}
\label{tab:spectra}
\renewcommand\arraystretch{1.2}
\small
\resizebox{0.7\linewidth}{!}{
\begin{tabular}{ccccc}
\Xhline{1pt}   
\multicolumn{1}{c}{Dataset}     & \multicolumn{1}{c}{Modality} & OA($\%$)	& AA($\%$)	& $\kappa$($\%$) \\ \hline
\multirow{2}{*}{Pavia Centre}   & RGB             & 82.85 & 55.96 & 75.27 \\
                                & HSI             & \textbf{90.62} & \textbf{69.63} & \textbf{86.75} \\ \hline
\multirow{2}{*}{Chikusei}       & RGB             & 79.95 & 80.54 & 74.09 \\
                                & HSI             & \textbf{88.33} & \textbf{85.11} & \textbf{84.98} \\ \hline
\multirow{2}{*}{AeroRIT}        & RGB             & 75.55 & 74.22 & 64.88 \\
                                & HSI             & \textbf{82.18} & \textbf{79.98} & \textbf{73.83} \\ \Xhline{1pt}   
\end{tabular}
}
\end{table}

The classification results of this RGB-based variant and our full HSI-based method are presented in Table~\ref{tab:spectra}. The comparison clearly shows that the use of full hyperspectral information leads to higher overall accuracy, higher average accuracy, and a better kappa coefficient on all datasets. This is expected, since hyperspectral data provide much richer spectral information than RGB images: each material or land-cover type tends to have a characteristic spectral curve across many bands, which is not fully captured by only three broad RGB channels. As a result, the classifier trained on HSI can more easily separate classes that appear similar in RGB but differ in their spectral signatures.

In addition, the spectral similarity among samples of the same class helps to reduce the negative effect of noisy labels. When training with hyperspectral data, mislabeled samples are more easily detected and corrected by the model because they are surrounded by many spectrally similar neighbors with correct labels. For example, if a pixel from class A is mistakenly assigned to class B in the pseudo-labels, its spectral neighbors from class A still form a tight cluster in the spectral space. During training, the model learns that these neighbors share similar spectral features and should be assigned to the same class, which provides indirect evidence to correct the mislabeling. In contrast, when only RGB information is available, the differences between some classes become much smaller, and this kind of spectral redundancy is lost, making the training process more sensitive to label noise. Overall, these results confirm that explicitly using hyperspectral information is important for robust classification in the presence of noisy pseudo-labels.

\subsubsection{Scaling Strategy}
We next study the influence of the resolution scaling (RS) strategy in the pseudo-label generation stage. As described in the main text, the RS strategy aims to better recognize objects of different sizes by fusing prediction results from multiple image resolutions. To evaluate its effect, we compare our full method with a variant that does not use RS, that is, pseudo-labels are generated at a single resolution and directly used for subsequent training. For reference, we also report the results of ClearCLIP and SegEarth-OV with and without the RS strategy under the same setting. The quantitative results are shown in Table~\ref{tab:rs}.

From the table, we can observe two main trends. First, the adoption of the RS strategy leads to a clear and consistent improvement for all three methods (ClearCLIP, SegEarth-OV, and our approach) across most metrics. This indicates that multi-scale fusion is a general and effective technique for enhancing CLIP-based dense prediction in remote sensing. Second, among all variants, our full method with RS achieves the best classification performance on all evaluated datasets, which further supports the effectiveness of combining RS with spectral noisy label learning.

These results are closely related to the fixed receptive field of CLIP models. Since the spatial resolution of CLIP feature maps is limited, a single input scale cannot handle both small and large objects equally well. Higher input resolutions are more suitable for recognizing small objects and thin structures (e.g., cars, narrow roads, small buildings), because these objects occupy more pixels and are less likely to vanish during feature extraction. On the other hand, lower resolutions are better at capturing global context and large homogeneous regions (e.g., water bodies, farmland, large roofs), since they force the model to focus on coarse structures and reduce the influence of local noise. By averaging prediction results from different resolutions, the RS strategy allows us to benefit from both types of information, leading to more reliable and stable pseudo-labels. These improved pseudo-labels then serve as a stronger supervision signal in the later noisy label learning stage.

\begin{table}[]
\centering
\caption{Ablation results on the scaling strategy.}
\label{tab:rs}
\renewcommand\arraystretch{1.1}
\small
\resizebox{0.95\linewidth}{!}{
\begin{tabular}{cccccc}
\Xhline{1pt} 
\multicolumn{1}{c}{Dataset}      & \multicolumn{1}{c}{Setting}    &    Method       & OA($\%$)       & AA($\%$)       & $\kappa$($\%$) \\ \hline
\multirow{6}{*}{Chikusei} & \multirow{3}{*}{w/o RS} & ClearCLIP   & 62.09          & 48.36          & 49.54          \\
                          &                         & SegEarth-OV & 75.54          & 59.96          & 67.49          \\
                          &                         & Ours        & \textbf{83.56} & \textbf{76.03} & \textbf{78.81} \\ \cline{2-6}
                          & \multirow{3}{*}{RS}     & ClearCLIP   & 72.14          & 57.76          & 62.64          \\
                          &                         & SegEarth-OV & 77.02          & 65.08          & 69.49          \\
                          &                         & Ours        & \textbf{88.33} & \textbf{85.11} & \textbf{84.98} \\ \hline
\multirow{6}{*}{Pavia Centre}   & \multirow{3}{*}{w/o RS} & ClearCLIP   & 87.02          & 53.28          & 81.33          \\
                          &                         & SegEarth-OV & 85.64          & 49.51          & 79.00          \\
                          &                         & Ours        & \textbf{88.33} & \textbf{58.48} & \textbf{83.43} \\ \cline{2-6}
                          & \multirow{3}{*}{RS}     & ClearCLIP   & 88.30          & 55.94          & 83.24          \\
                          &                         & SegEarth-OV & 88.24          & 55.79          & 82.91          \\
                          &                         & Ours        & \textbf{90.62} & \textbf{69.63} & \textbf{86.75} \\ \hline
\multirow{6}{*}{AeroRIT}  & \multirow{3}{*}{w/o RS} & ClearCLIP   & 69.47          & 64.82          & 54.28          \\
                          &                         & SegEarth-OV & 70.34          & 58.67          & 54.50          \\
                          &                         & Ours        & \textbf{80.11} & \textbf{78.81} & \textbf{71.10} \\ \cline{2-6}
                          & \multirow{3}{*}{RS}     & ClearCLIP   & 74.52          & 62.48          & 60.65          \\
                          &                         & SegEarth-OV & 73.66          & 58.93          & 59.07          \\
                          &                         & Ours        & \textbf{82.18} & \textbf{79.98} & \textbf{73.83} \\ \Xhline{1pt} 
\end{tabular}
}
\end{table}

\subsubsection{Label Refinement Strategy}
Finally, we analyze the effect of the proposed label refinement strategy, which relies on random, confident, and hard sets, as well as soft labels obtained from Gaussian Mixture Models. We conduct ablation experiments on the Pavia Centre dataset and report the results in Table~\ref{tab:subset}. In these experiments, we gradually add different subsets into the training process to see how each component contributes to the final performance.

When we train the spectral classifier using only the random set with hard labels (i.e., the pseudo-labels given by SegEarth-OV), the model already achieves a reasonable level of performance, because the random set distills the coarse semantic prior from the PLG stage. However, the presence of label noise still limits the classification accuracy, especially for classes with strong confusion or limited training samples. When we include the confident set with soft labels in the training, the performance improves noticeably. The confident set contains pixels whose predictions are both consistent and reliable according to the BvSB criterion and the GMM separation. The use of soft labels for these samples provides more flexible supervision than one-hot labels, allowing the model to better adapt to subtle class boundaries while still focusing on high-confidence information.

The inclusion of the hard set brings a further, although sometimes smaller, gain. Samples in the hard set are more likely to be mislabeled, but they often lie near decision boundaries or in mixed regions, and thus carry important information about class overlap and transition zones. By assigning soft labels to these samples and using a suitable weight in the loss function, we allow the model to explore these difficult regions without overfitting to incorrect labels. This helps the model to refine its decision boundaries and improve the classification of ambiguous pixels. The overall results in Table~\ref{tab:subset} show that the addition of training samples with soft labels from both the confident and hard sets significantly improves performance compared to using only the random set.

\begin{table}[t]
\centering
\caption{Ablation results on the training subsets, where R, C, and H denote the random, confident, and hard sets, respectively.}
\label{tab:subset}
\renewcommand\arraystretch{1.1}
\resizebox{0.6\linewidth}{!}{
\begin{tabular}{ccc|ccc}
\Xhline{1pt} 
R & C & H & OA($\%$)       & AA($\%$)       & $\kappa$($\%$) \\ \hline
\ding{51}          &               &          & 78.68          & 63.62          & 71.08          \\
\ding{51}          & \ding{51}     &          & 81.21          & 64.81          & 74.30          \\
\ding{51}          &               & \ding{51}& 69.35          & 60.22          & 59.77          \\
\ding{51}          & \ding{51}     & \ding{51}& \textbf{90.62} & \textbf{69.63} & \textbf{86.75} \\ \Xhline{1pt} 
\end{tabular}
}
\end{table}

\section{Conclusion}\label{sec:conclusion}
In this paper, we propose a novel zero-shot HSI classification framework, SPECIAL, which leverages the capabilities of CLIP for pixel-level classification without the need for manual annotation. The framework consists of two stages, including CLIP-based pseudo-label generation and noisy label learning. In the pseudo-label generation stage, the framework interpolates HSI data to obtain RGB bands and uses CLIP for initial classification, generating noisy pseudo-labels and confidence scores. A scaling strategy is proposed to improve the quality of the pseudo-labels. In the noisy label learning stage, spectral information and a label refinement strategy are incorporated to alleviate the issue of label noise and further enhance accuracy. Experiments on three datasets demonstrate SPECIAL's superiority over existing approaches in zero-shot HSI classification.

\bibliographystyle{IEEEtran}  
\bibliography{ref}     

\vfill

\end{document}